\documentclass[journal]{IEEEtran}
\usepackage{cite}
\usepackage{amsmath,amssymb,amsfonts}
\usepackage{algorithmic}
\usepackage{graphicx}
\usepackage{textcomp}

\usepackage{dblfloatfix}    % To enable figures at the bottom of page

\usepackage[table, dvipsnames]{xcolor}
\newsavebox{\measurebox}

\usepackage{cite}
\usepackage{amsmath,amssymb,amsfonts}
\usepackage{algorithmic}
\usepackage{graphicx}
\usepackage{textcomp}
\usepackage{subfig}
\usepackage{color}
\usepackage{graphicx}
\usepackage{afterpage}
\usepackage{cite}
\usepackage{url}
\usepackage{placeins}
\usepackage{pdflscape}
\graphicspath{{}}
\usepackage{textcomp}
\usepackage{notoccite}
\usepackage{makecell}
\usepackage{multirow}
\usepackage{longtable}
\usepackage{booktabs}
\usepackage{booktabs}
\usepackage{array}
\usepackage{pdflscape}
\usepackage{amsmath}
\usepackage{mathtools}
\usepackage{amsfonts}
\usepackage{hyperref}
\usepackage{graphicx}

\usepackage{url}
\usepackage{placeins}
\usepackage{pdflscape}
\usepackage{multirow}
\usepackage{rotating}
\usepackage[mathlines]{lineno}
\usepackage{color}

\usepackage{lineno,hyperref}
\newcommand{\cmark}{\ding{51}}%
\newcommand{\xmark}{\ding{55}}%
\usepackage{textcomp}

\usepackage{notoccite}
\usepackage{afterpage}
\usepackage{makecell}
\usepackage{multirow}
\usepackage{longtable}
\usepackage{amssymb}% http://ctan.org/pkg/amssymb
\usepackage{pifont}% http://ctan.org/pkg/pifont
\def\BibTeX{{\rm B\kern-.05em{\sc i\kern-.025em b}\kern-.08em
		T\kern-.1667em\lower.7ex\hbox{E}\kern-.125emX}}

\ifCLASSINFOpdf
\else
\fi
\begin{document}
\title{Deep Neural Networks Based Weight Approximation and Computation Reuse for 2-D Image Classification}

\author{Mohammed F. Tolba, Huruy Tekle Tesfai,
	Hani  Saleh,~\IEEEmembership{Senior Member, IEEE,}
	Baker Mohammad ,~\IEEEmembership{Senior Member, IEEE,}
	and~Mahmoud Al-Qutayri,~\IEEEmembership{Senior Member, IEEE}% <-this % stops a space
	\thanks{Mohammed F. Tolba, Huruy Tekle Tesfai, Hani  Saleh, ~Baker Mohammad  and Mahmoud Al-Qutayri  are with System-on-Chip (SoC) Center, Khalifa University, Abu Dhabi, UAE, P.O. Box 127788, Abu Dhabi, UAE.}% <-this % stops a space
	\thanks{Manuscript received April 19, 2005; revised August 26, 2015.}
}


\markboth{Journal of \LaTeX\ Class Files,~Vol.~14, No.~8, August~2015}%
{Shell \MakeLowercase{\textit{et al.}}: Bare Demo of IEEEtran.cls for IEEE Journals}



\maketitle

\begin{abstract}
	
	Deep Neural Networks (DNNs) are computationally and memory intensive, which makes their hardware implementation a  challenging task especially for resource constrained devices such as IoT nodes. \color{black} To address this challenge, this paper introduces a new method to improve DNNs performance by fusing approximate computing with data reuse techniques to be used for image recognition applications. \color{black} DNNs weights are approximated based on the linear and quadratic approximation methods during the training phase, then, all of the weights are replaced with the linear/quadratic coefficients to execute the inference in a way where different weights could be computed using the same coefficients. This leads to a repetition of the weights across the processing element (PE) array, which in turn enables the reuse of the DNN sub-computations (computational reuse) and leverage the same data (data reuse) to reduce DNNs computations, memory accesses, and improve energy efficiency albeit  at the cost of increased  training time.  \color{black} Complete analysis for both MNIST and CIFAR 10 datasets is presented for  image  recognition \color{black}, where LeNet 5 revealed a reduction in the number of parameters by a factor of $1211.3\times$  with a drop of less than 0.9\% in accuracy. When compared to the state of the art Row Stationary (RS) method, the proposed architecture saved $54\%$ of the total number of  adders and multipliers needed. Overall, the proposed approach is suitable for IoT edge devices as it reduces the memory size requirement as well as the number of needed memory accesses.

	%   Using this repetition to reuse DNN sub-computations and to reduce DNNs memory storage.
	%On CNN MNIST dataset,
	%The proposed algorithm  achieve a reduction rate of $2.5\times$  for each convolution layer, and $48\times$  for each FC layer without any loss of accuracy for CNN architecture.
	%  The proposed technique makes the memory storage so small that all weights can be stored on-chip instead of going to off-chip energy-consuming DRAM    which makes the use of the proposed technique more suitable for IoT applications. Proposed method  facilitates the use of complex neural networks in mobile applications where application size and download bandwidth are constrained.
\end{abstract}


% \fi
\IEEEpeerreviewmaketitle

\section{Introduction}
\label{sec:introduction}
\IEEEPARstart{D}{eep}  neural networks (DNNs) are the backbone of  artificial intelligence (AI) and are used in a wide range of applications. Recently, DNNs started to extend to Internet-of-Things applications \cite{sun2018resinnet,lecun2015deep,gubbi2013internet,guo2018mobile},  as well as
other major applications of DNNs including  computer vision, speech recognition, robotics, autonomous vehicles \cite{chen2015deepdriving}, cancer detection \cite{esteva2017dermatologist} and video games \cite{silver2016mastering}. In the IoT domain, deep learning techniques can further be used to extract user behavior and ambient context from sensor data  given its strong ability to learn abstract representations \cite{lane2015early}. The projection of fifty Billion Internet-of-Things (IoT) devices by 2020 has pushed the energy harvesting research to investigate new solutions for improving the lifetime of batteries \cite{alhawari2016efficient,alhawari2016power}.   Developing  efficient designs and hardware computational 
architectures for DNNs are attracting the attention of researchers due to the many challenges that need to be overcome.   
% Recently, DNNs were able to achieve superior accuracy compared with other machine learning methods. 
% Yet, this leads to new challenges for existing hardware systems. Traditional general-purpose processors do not provide enough performance and energy efficiency, particularly for real-time applications.  
%The situation gets even more demanding when AI algorithms need to run on the edge of smart mobile devices for example.
The challenges are particularly critical for DNNs implementation on resource-constrained devices such as IoT nodes. 
%For various mobile-first tech giants like Facebook and Baidu, software programs get upgraded from various app stores, and they are sensitive to the size of the files that get downloaded \cite{han2015deep}. For example, App Store has the limitation “apps above 100 MB will not download until you connect to Wi-Fi”. Therefore, a feature that increases the binary size by 100MB will get much more scrutiny than one
%that raises it by 10MB. 
Moreover, having DNNs running on edge and mobile  devices have lots of advantages such as security, bandwidth requirement, and real-time  processing. However the memory overhead limits DNNs from being incorporated into  edge and mobile devices  on a large scale.
Advanced DNNs need tens to hundreds of megabytes of parameters and billions of computations for a particular inference \cite{horowitz20141}. This results in long inference time and even longer training time.   DNNs extensively use  lots of multiply and accumulate (MAC) operations in the convolution and FC (fully connected)  layers, which account for over 99\% of the overall computations \cite{yang2017designing}.  Large storage memory, rapid data movement and an enormous number of computations make it very challenging   to directly implement DNNs based AI applications on existing edge devices hardware due to their limited computing and power resources.
\begin{figure}[t!]
	\centering
	{\includegraphics[width=0.85\linewidth]{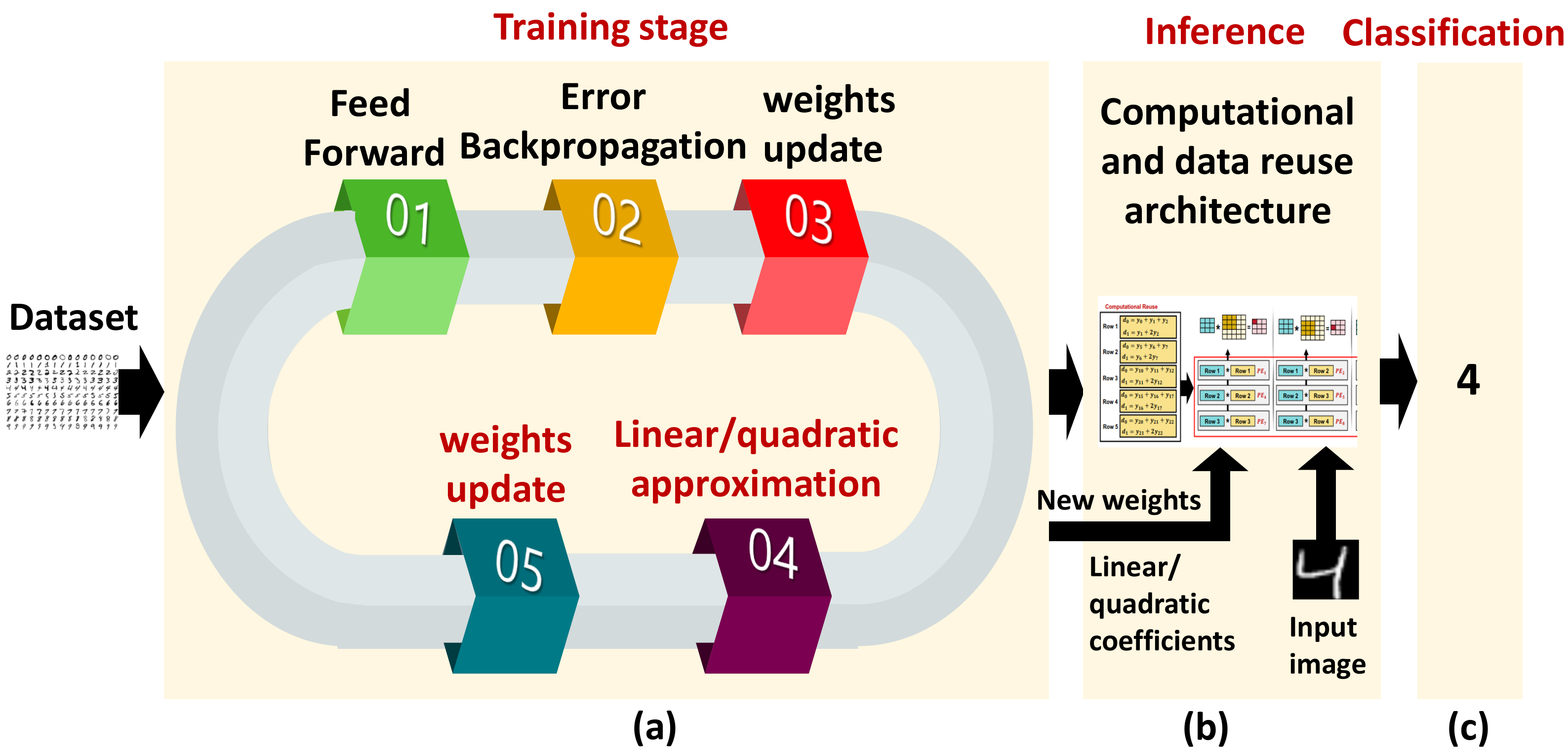}
		\caption{Block level description of major blocks including proposed approach. (a) the training stage, linear and quadratic approximation techniques are used to approximate the DNN weights, (b) Inference stage,  and (c) classification.   }
		\label{fig}}
\end{figure}

\begin{table*}[]
	\centering
	\caption {\color{black}The contribution of different approximation techniques to DNN inference acceleration in custom hardware \cite{wang2019deep}.}
	\label{Tab00}
	\begin{tabular}{|c|c|c|c|c|}
		\hline
		\multicolumn{2}{|c|}{Method}                                        & \begin{tabular}[c]{@{}c@{}}Cheaper arithmetic operations\end{tabular} & \begin{tabular}[c]{@{}c@{}}Memory reduction\end{tabular} & \begin{tabular}[c]{@{}c@{}}Workload reduction\end{tabular} \\ \hline
		\multirow{4}{*}{	\begin{tabular}{@{}c@{}}{}	Quantization \end{tabular} }      & \begin{tabular}[c]{@{}c@{}}Fixed-point representation   \end{tabular}  & \cmark                                                                       & \cmark                                                          & \xmark                                                            \\ \cline{2-5} 
		&\begin{tabular}[c]{@{}c@{}} Binarisation and ternarisation \end{tabular} & \cmark                                                                       & \cmark                                                          & \xmark                                                            \\ \cline{2-5} 
		& Logarithmic quantisation       & \cmark                                                                       & \cmark                                                          & \cmark  (if shift lengths are constant)                                                          \\ \hline
		\multirow{5}{*}{\begin{tabular}[c]{@{}c@{}}	Weights  reduction\end{tabular}} & Pruning                        & \xmark                                                                      & \cmark                                                          & \cmark                                                            \\ \cline{2-5} 
		& Weight sharing                 & \xmark                                                                      & \cmark                                                         & \cmark (If multiplication are precomputed)                                                           \\ \cline{2-5} 
		& Low-rank factorisation         & \xmark                                                                       & \cmark                                                         & \cmark                                                            \\ \cline{2-5} 
		& Structured matrices            & \xmark                                                                       & \cmark                                                          & \cmark                                                           \\ \cline{2-5} 
		& Knowledge distillation         & \xmark                                                                       & \cmark                                                         & \cmark                                                            \\ \hline
	\end{tabular}
\end{table*}  \color{black}
\color{black} The  main goal of this work is to reduce the number of computations and memory storage for DNNs inference so they can be particularly  used  for image  recognition  on  edge  resource-constrained devices, such as IoT nodes. \color{black}
The idea is to merge data reuse and approximate computing techniques to develop computational reuse method. In this method a specific computation operation is reused multiple times in the PE array processor. 
The computational reuse method relies on the repetition of a specific number of computations inside the PE array. Based on the idea of taking a common factor of a specific number of computations (DNNs computations), a reduction of DNNs computations is achieved. 
Block level description of the proposed approach is shown in Fig. \ref{fig}, where  linear and quadratic approximation techniques are used to approximate  DNNs weights in the back-propagation during the training stage. Then, all weights are replaced with the linear/quadratic coefficients to implement the inference using the proposed data/computational reuse architecture. Such that different weights could be computed using the same coefficients.
This leads to a sizable  repetition of the weights across the PE array.  Based on this repetition, DNN sub-computations   are reused to reduce DNNs memory storage and improve energy efficiency albeit  at the cost of increased the training time.  In addition to the computational reuse, data reuse method is proposed.

%Adding more operations like te approximation in the backpropagation during training makes it further expensive and demands a longer time. 
%However, the approximation of DNNs weights reduces the number of parameters, computations, and memory accesses in the inference part.
\color{black}
The
main contributions of this paper can be summarized as follows:
We proposed a new algorithm to optimize the training of DNNs to reduce the number of DNNs parameters substantially by approximating the DNNs weights using linear and quadratic approximations.  	Using the proposed training method leads to a reduction of the DNN weights and a repetition of the weights in the DNN, which in turn enables the reuse of the DNN sub-computations (computational reuse) and leverage the same data (data reuse) to reduce DNNs computations, memory storage, and energy, which makes it  suitable for IoT applications.
Complete analysis for both MNIST and CIFAR 10 datasets is presented for image recognition, where LeNet 5 revealed a reduction in the number of parameters by a factor of $1211.3 \times$    in the inference phase with a drop of less than 0.9\% in accuracy. The results show the effectiveness of our method and distinct advantages over other  methods  reported in the literature. 
When compared to the state of the art Row Stationary (RS) method, the proposed architecture saved 54\% of the total number of adders and multipliers needed. 

\color{black}
%A new algorithm is introduced to reduce the number of DNNs parameters substantially by approximating the DNNs weights using   linear and quadratic approximations methods.  
%Applying the proposed approximation method in the training results in a reduction rate of $47.8 \times$  on LeNet 300-100 parameters and $40.4\times$ on LeNet 300-100  computations in the inference phase.  
%In the inference phase, a computational reuse technique is proposed to reduce the number of MAC operations inside the PE array. 
%The computational and parameters proposed techniques resulted in reduced memory storage requirements and less overall memory accesses. 	

The rest of the paper is organized as follows.  Section \ref{related_work} presents previous related works on DNN techniques.  Section \ref{algorithm} introduces the proposed  approximation algorithm.   Section \ref{ex:sec} shows different experiments to verify the proposed  approximation algorithm. Section \ref{reuse} demonstrates the  opportunity for  data reuse and computation reuse.  Finally, Section \ref{sec:conclusion} concludes the paper.

% \usepackage{multirow}

\section{Related Work} \label{related_work}

Training DNNs  is an expensive task requiring specialized hardware such as GPUs \cite{wang2019accelerated}.  Hundreds of millions of parameters in the DNNs need to be iteratively updated hundreds of thousands of times during the training phase.  Although training demands more computations when
compared to inference, it is necessary to decrease the cost
of inference as much as possible, because it is the inference
that is normally subject to stricter  real-world design
constraints \cite{kim2020effects}.  \color{black}
Approximating the DNNs weights during the training cause additional computations which lead to longer training time. However, the approximation of DNNs weights reduces the number of parameters, resulting in  higher parallelism, lower off-chip memory transfer,  improving the energy efficiencies and power consumptions \cite{wang2019deep} and \cite{guo2017new}.
\color{black}

An important characteristic of DNNs parameters is their error-resilience, which can be exploited by using the fundamentals of approximate computing to design energy-efficient hardware accelerators \cite{reda2019approximate}. Based on this, various  researchers successfully employed approximate computing techniques to DNNs. \color{black} The state-of-the-art methods use approximate computing to improve the energy efficiency and performance of DNN hardware through logarithmic representation, pruning, quantization, weight sharing … etc \cite{reda2019approximate,9154572,9381660}.   However, using approximate computing in DNNs results in the need for retraining the network to recover the accuracy loss due to using approximate methods. 
Based on logarithmic representation, DNNs parameters are quantized
into powers of two with a scaling factor \cite{wang2019deep}.  The range of the weights is more valuable than its precision in saving network accuracy. Logarithmic representations can display the DNNs weights over a very wide range of values using a few bits \cite{lai2017deep}. The weight reduction method enhances the performance of hardware inference by decreasing both workload and off-chip memory.
Pruning is an iterative technique to prune weights of small values and retrain the DNN \cite{guo2017new}. Similar neurons can be wired together and therefore pruned away.
Pruning accomplished 9× reduction in the number of weights on AlexNet for ImageNet dataset with minimum accuracy degradation \cite{han2015deep}. 
Weight sharing arrange parameters into groups, decreasing network
size and using cheaper multiplications.  Weight sharing can be
done in different steps \cite{han2015deep}. The network is firstly pruned \cite{han2015learning}, then weights are quantized through k-means clustering.  After that, the quantized network is retrained again. Finally, the quantized weights are compressed with Huffman coding to  decrease the memory storage of neural networks without affecting their accuracy. Table \ref{Tab00} illustrates the contribution of different approximation techniques to DNN inference acceleration in custom hardware \cite{wang2019deep}. The quantization techniques improve parallelism by the use of cheaper arithmetic units. Weight-reduction techniques decrease the number of parameters, saving memory, and reducing the workload. \color{black}
DNNs quantization and weight sharing methods are used to reduce the number of bits needed for each weight. 
Previous work discussed opportunities that leverage weight reuse to reduce the energy efficiency of CNN, based on refactoring and reusing CNN sub-computations \cite{hegde2018ucnn}. 

The data access can be minimized by reusing the same piece of data as much as possible. The problem is that the storage capacity of low-cost memories is limited. Different dataflows methods have been explored to maximize data reuse \cite{chen2016eyeriss,sze2017efficient}. Some of the well-known data flow methods include weight stationary (WS) \cite{chakradhar2010dynamically,gokhale2014240}, output stationary (OS) \cite{peemen2013memory} and row stationary (RS) \cite{sze2017efficient} dataflows. WS data flow is used to reduce the energy consumption of accessing the weights through maximizing the weights accesses from the register file (RF) at the PE.  The OS data flow reduces the energy consumption of accessing the partial sums. The RS dataflow increases the reuse at the RF for all kinds of data to achieve overall energy efficiency \cite{chen2016eyeriss}. 

\begin{figure}[]
	\centering	
	{\includegraphics[width=0.85\linewidth]{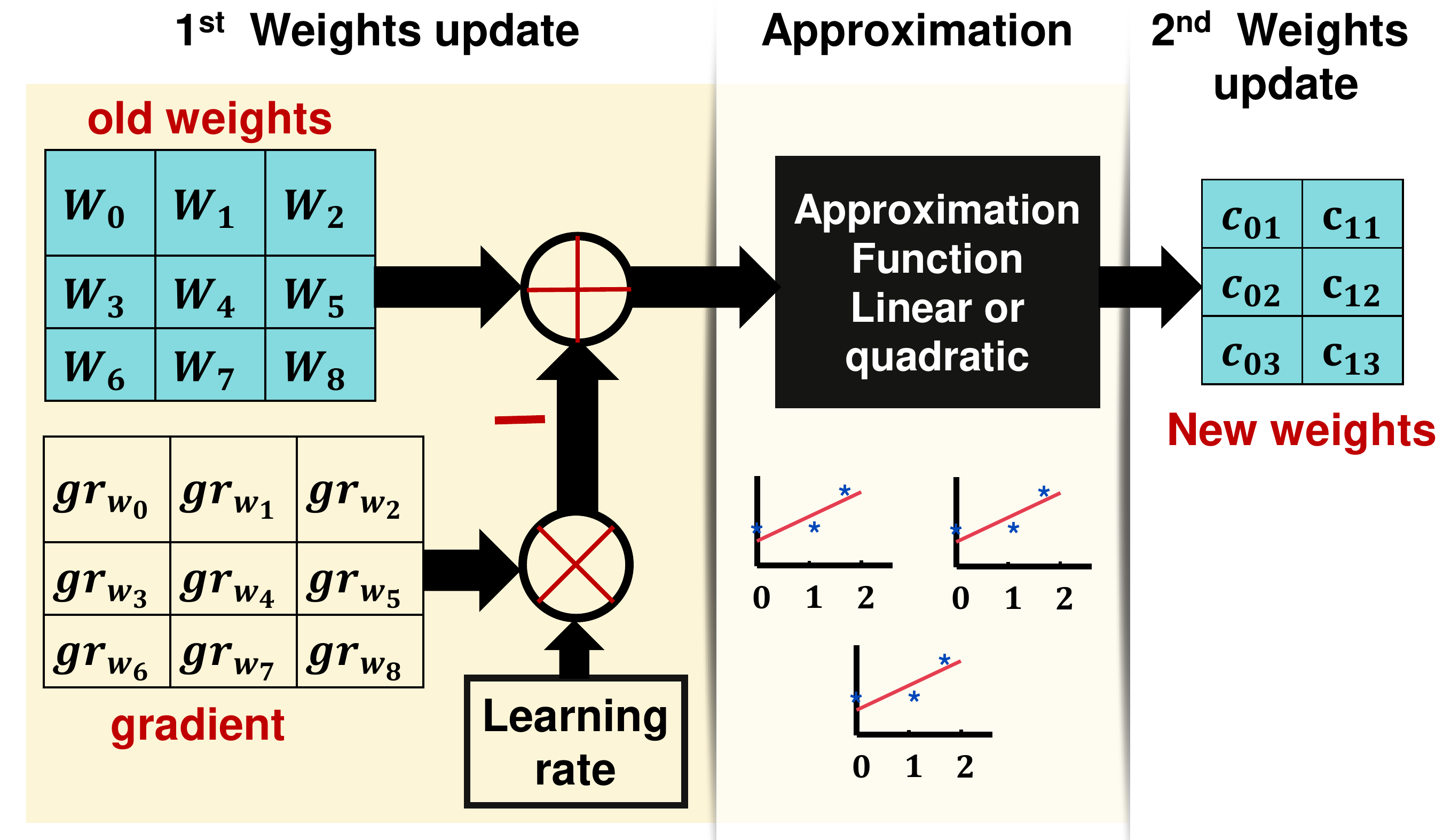}
		\caption{Proposed linear/quadratic approximation algorithm in the training phase, where all the weights are approximated for each batch during the back propagation.    }
		\label{fig_7}}
\end{figure}

\begin{figure*}[t!]
	\centering
	{\includegraphics[width=0.8\linewidth]{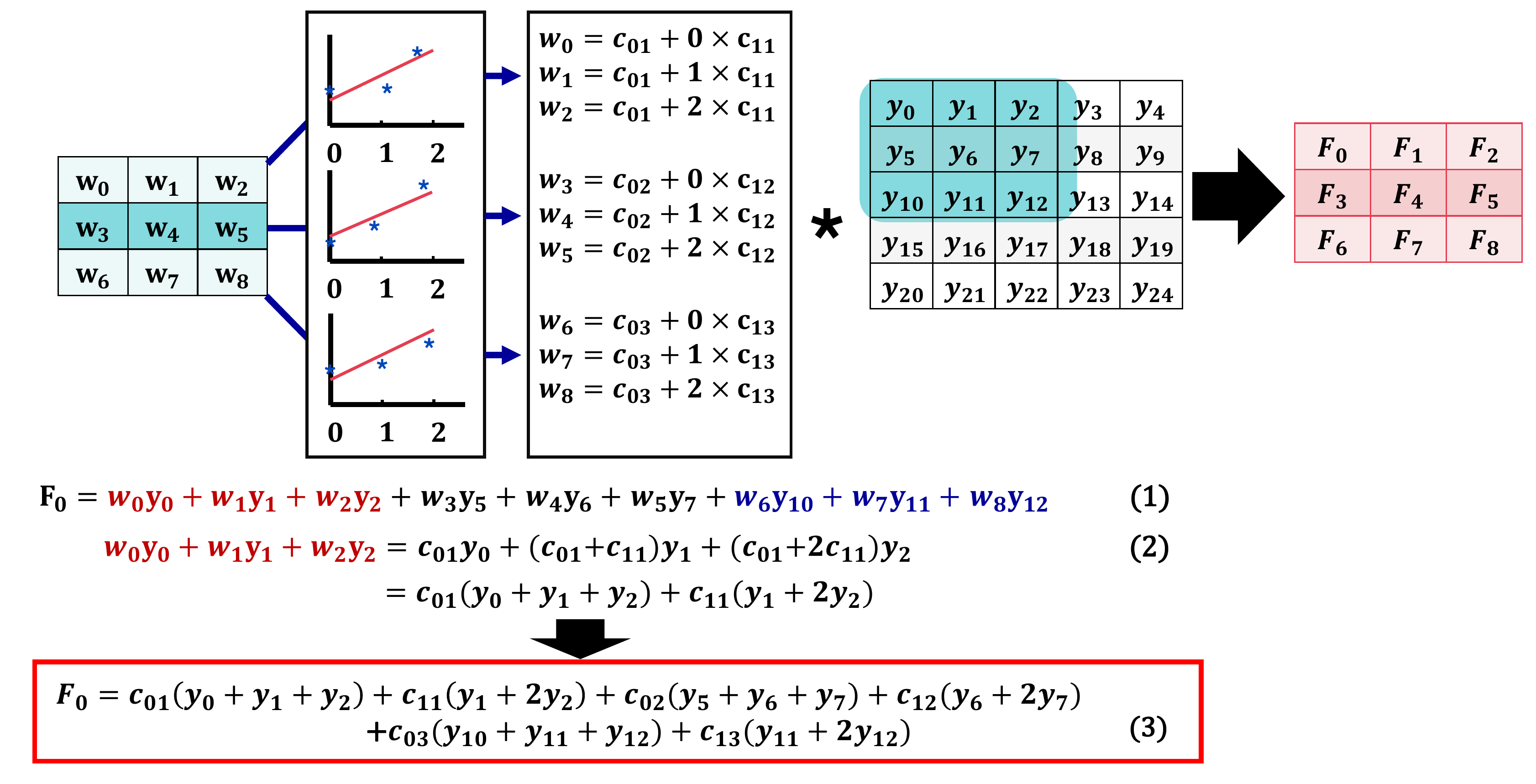}
		\caption{Proposed linear approximation algorithm used to reduce CNN weights, this example for $3\times3$ filter and $5\times5$ input feature map ($ifmap$). Each filter row is approximated linearly to produce the approximated weights.  }
		\label{fig_1}}
\end{figure*}

\section{ Proposed  approximation algorithm} \label{algorithm}
Not all the DNNs weights have to be retained at the exact value achieved by training, many of these weights can be approximated using a number of approximation methods. 
In this section, we introduced a method  to approximate the DNNs weights during the training as follows:
step 1, in the Backpropagation: 
after updating all DNNs weights, all the weights are divided into a uniform group of weights.
Then, each group is approximated using linear/quadratic approximation methods.
For each group of weights, linear/quadratic coefficients  are  used to compute the new DNNs weights.
Step 2, in the  feed forward: the new weights are used to perform the feed forward path.
The previous steps are performed for each batch and epoch during the training phase. 

Figure \ref{fig_7} presents the proposed linear/quadratic approximation in the training phase.  On the top left is the $3 \times 3$ weight matrix, and on the bottom left is the $3 \times 3$ gradient matrix. All the gradients are multiplied by the learning rate and subtracted from the old weights ($1^{st}$ weights update part). After that, all the updated weights are linearly or quadratically approximated  to computes the new weights or linear/quadratic equation coefficients. This approximation process is applied in the back-propagation for each batch. Using the MATLAB polyfit function to approximate the weights required a long time in the training. As the X-axis is fixed in our approximation during the training, the function can be improved by performing some of the computations in the first iteration and then the results can be reused in all other iterations to reduce the training time.

\begin{figure*}[]
	\centering
	{\includegraphics[width=0.85\linewidth]{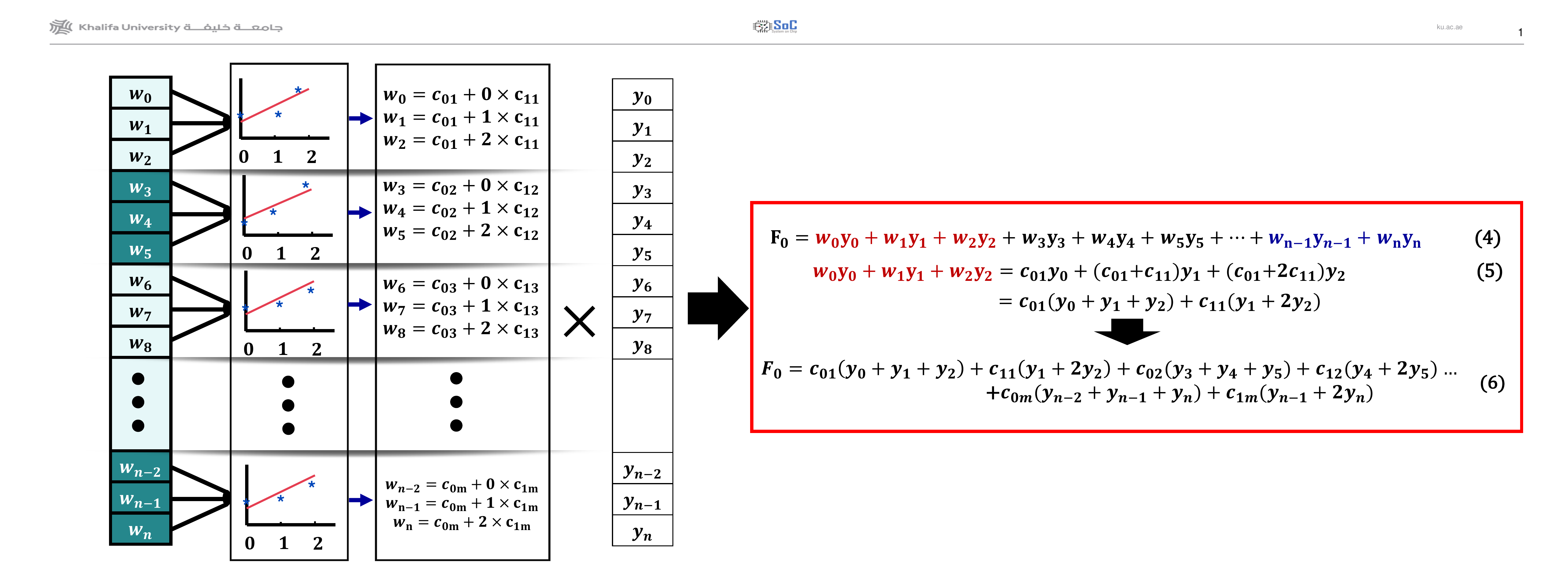}
		\caption{Proposed linear approximation algorithm used to reduce FC weights. All the FC weights are divided to uniform groups each one has three weights. Each group is approximated linearly to get the approximated weights.  }
		\label{fig_4}}
\end{figure*} 
\begin{figure*}[t!]
	\centering
	{\includegraphics[width=0.85\linewidth]{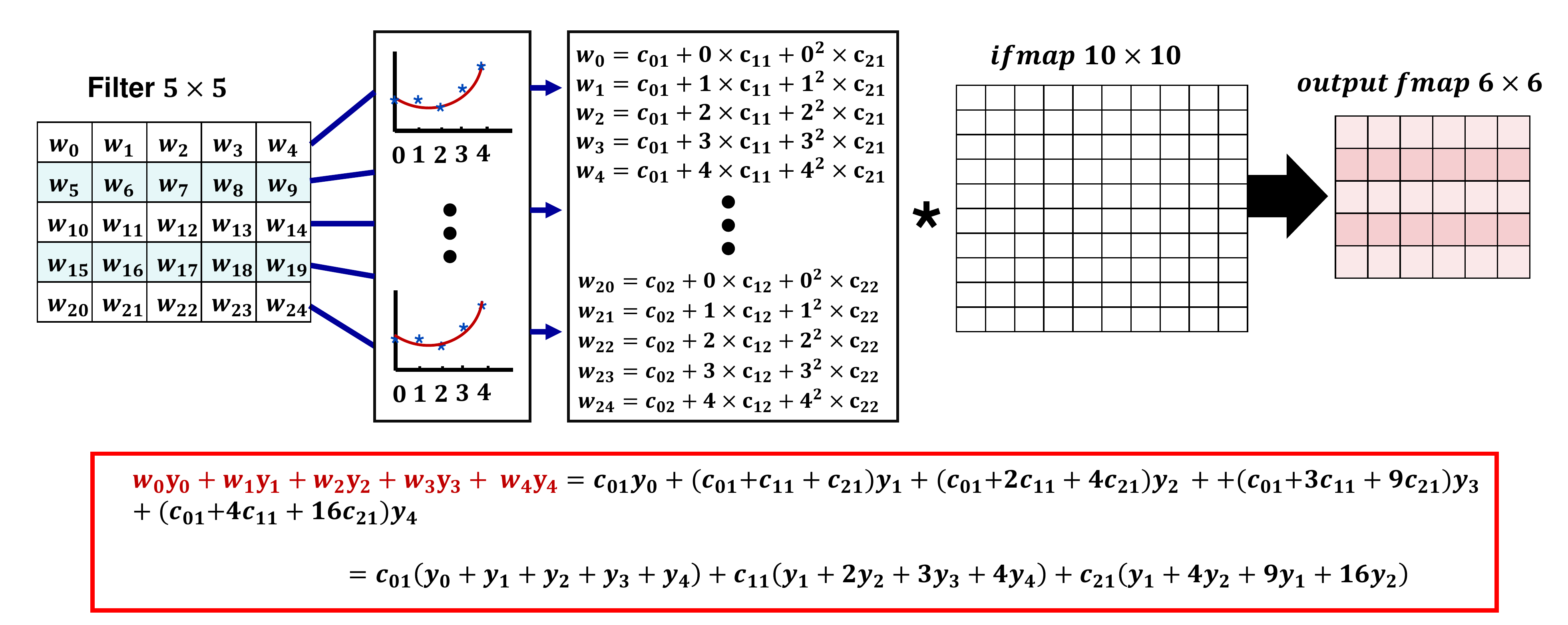}
		\caption{Proposed quadratic approximation algorithm used to approximate CNN weights, this example for $5\times5$ filter and $10\times10$ input feature map ($ifmap$). Each filter row is approximated using a $2^{nd}$ order polynomial to produce the approximated weights.  }
		\label{fig_2nd}}
\end{figure*}

\subsection{Linear approximation method}
Figure \ref{fig_1} presents an example that describes the proposed algorithm for approximating   CNNs weights.  In this example, the input image size is $5\times5$ and the filter size is $3\times3$. The linear  approximation method is applied for the filter  in which any filter value could be computed using $c_{0j}+xc_{1j}$, where $c_{0j}$ is the first coefficient, $c_{1j}$ depicts the second coefficient, $x$ could be $0, 1, 2$ and $j$ represents the filter row number. Generally, the linear approximation can be applied as follows:
Each filter row is taken as a group of weights and each group  is approximated using the linear approximation method in interval $[x=0:N_w]$, where $N_w$ depicts the number of weights in each row.
The   coefficients of the linear equation in each row  are then used to compute the new  weights for each filter.
Each filter value is replaced with its approximate value that is computed using the  coefficients of the linear equation as shown in Fig. \ref{fig_1}.

As can be seen, the convolution outputs could be computed traditionally as shown in Fig. \ref{fig_1} equation (1)  or using the proposed method in equation (3).  The number of multipliers and weights  are reduced by a third compare with the traditional method. But, the number of adders is increased from 8 to 14. However, the number of adders could be reduced by reusing the same computation operation multiple times in the PE array processor (proposed computation reuse technique). 
For  FC layer, all the weights are divided to uniform groups and each group of weights is linearly approximated as shown in Fig. \ref{fig_4}.  
In this figure, the input image size is $n\times1$ and the filter size is $n\times1$. Similarly to CNN approximation, any filter value could be computed using $c_{0j}+xc_{1j}$.

\subsection{Quadratic approximation method}
Increasing the number of weights inside the linear interval also increases the reduction rate but reduces accuracy. To enhance accuracy, a quadratic approximation method is used to approximate the DNNs weights along with the  linear approximation method presented in the previous subsection, the quadratic method can be used to approximate the CNN  weights as presented in the example in Fig. \ref{fig_2nd}. 
In this example, the input image size is $10\times10$ and the filter size is $5\times5$.  The quadratic  approximation method is applied for the filter  in which any filter value could be computed using $c_{0j}+xc_{1j}+x^2c_{2j}$, where $c_{0j}, c_{1j}$ and $c_{2j}$ are the binomial coefficients, $x$ could be $0, 1, 2, 3, 4$ and $j$ represents the filter row number. Each row can be computed as shown in the equation presented in Fig. \ref{fig_2nd}.  The FC layer can also be approximated using quadratic method like CNN, which  all the weights are divided to uniform groups and each group of weights is  approximated  using a quadratic approximation method.

%\begin{table}[t!]
%	\centering
%	\caption {Testing accuracy  for different number of weights inside the linear interval  for MNIST FC with two hidden layers (64 and 32). $N_w$ represents the number of weights inside the linear interval.  All of these results are based on linear approximation method. The training is applied for 200 epochs.  }
%	\label{Tab1}
%	\begin{tabular}{|c|c|c|c|c|c|}
%		\hline
%		Approximation     &	\begin{tabular}[c]{@{}c@{}} Without \end{tabular} & {	\begin{tabular}[c]{@{}c@{}} During \\inference\end{tabular} } & \multicolumn{3}{c|}{	\begin{tabular}[c]{@{}c@{}} During training\end{tabular} }         \\ \hline
%		\begin{tabular}[c]{@{}c@{}}	$N_w$ \end{tabular}& 0       & 3                                & 3     &   4    & 8       \\ \hline
%		Accuracy                               & 97.53\%   & 31.78\%                    & 97.53\% & 97.53\% & 97.49\%  \\ \hline
%	\end{tabular}
%\end{table}

\begin{table*}[]
	\centering
	\caption {\color{black}Comparison with LeNet 300-100 (Fully connected approach) and  \cite{han2015deep,han2015learning} P: pruning, Q:quantization, H:Huffman coding, $N_p$  is the total number of parameters and $N_w$ depicts the number of weights inside the linear/quadratic interval. The linear and quadratic approximation are done during the training. Each weight needs 8-bits fixed point for proposed method. The training is applied for 200 epochs. KB depicts kilobyte. \color{black} }
	\label{Tab11}
	\begin{tabular}{|c|c|c|c|c|c|c|c|c|c|c|c|c|c|}
		\hline
		\multicolumn{2}{|c|}{}                                                                                                                                &                                                                         &                               &                               &                               & \multicolumn{8}{c|}{Proposed (64-32) linear and quadratic approximation during the training}                                                                                                                                                                                                                                                                                                                                                                                                  \\ \cline{7-14} 
		\multicolumn{2}{|c|}{}                                                                                                                                &                                                                         &                               &                               &                               & Without                        & \multicolumn{4}{c|}{Linear}                                                                                                                                                                                                & \multicolumn{3}{c|}{Quadratic}                                                                                                                                           \\ \cline{7-14} 
		\multicolumn{2}{|c|}{\multirow{-3}{*}{Method}}                                                                                                        & \multirow{-3}{*}{\begin{tabular}[c]{@{}c@{}}Ref\\ 300-100\end{tabular}} & \multirow{-3}{*}{\begin{tabular}[c]{@{}c@{}}P\\ \cite{han2015deep,han2015learning}\end{tabular} }           & \multirow{-3}{*}{\begin{tabular}[c]{@{}c@{}}P+Q\\ \cite{han2015deep}\end{tabular} }          & \multirow{-3}{*}{\begin{tabular}[c]{@{}c@{}}P+Q+H \\ \cite{han2015deep}\end{tabular} }       & Case 0                         & \begin{tabular}[c]{@{}c@{}}Case 1\end{tabular} & \begin{tabular}[c]{@{}c@{}}Case 2\end{tabular} & \begin{tabular}[c]{@{}c@{}} \textbf{Case 3} \end{tabular} & \begin{tabular}[c]{@{}c@{}}Case 4\end{tabular} & \begin{tabular}[c]{@{}c@{}}Case 5\end{tabular} & \begin{tabular}[c]{@{}c@{}}Case 6\end{tabular} & \begin{tabular}[c]{@{}c@{}}Case 7\end{tabular} \\ \hline
		& fc1                          & 235.2k                                                                  & -                             & -                             & -                             & 50.18k                         & 12.5k                                              & 6.3k                                                &  \textbf{ 4.1k}                                                   & 3.6k                                                   & 5.4k                                                  & 5.4k                                                    & 4.8k                                                   \\ \cline{2-14} 
		& fc2                          & 30k                                                                     & -                             & -                             & -                             & 2k                             & 0.5k                                               & 0.25k                                               &        \textbf{1k }                                                   & 1k                                                     & 0.75k                                                 & 0.4k                                                    & 0.4k                                                   \\ \cline{2-14} 
		& fc3                          & 1k                                                                      & -                             & -                             & -                             & 0.32k                          & 0.08k                                              & 0.32k                                               &   \textbf{0.32k}                                                & 0.32k                                                  & 0.32k                                                 & 0.32k                                                   &                                                        \\ \cline{2-14} 
		\multirow{-4}{*}{\begin{tabular}{@{}c@{}}\rotatebox[origin=c]{0}{$N_p$}\end{tabular}}                                                                                           & {\color[HTML]{9A0000} Total} & {\color[HTML]{9A0000} 266.2k}                                           & {\color[HTML]{9A0000} 22k}    & {\color[HTML]{9A0000} -}      & {\color[HTML]{9A0000} -}      & {\color[HTML]{9A0000} 52.5k}   & {\color[HTML]{9A0000} 13k}                         & {\color[HTML]{9A0000} 7k}                           & {\color[HTML]{9A0000} \textbf{5.5k}}                            & {\color[HTML]{9A0000} 4.9k}                            & {\color[HTML]{9A0000} 6.5k}                           & {\color[HTML]{9A0000} 6.1k}                             & {\color[HTML]{9A0000} 5.5k}                            \\ \hline
		& fc1                          &                                                                         &                               &                               &                               & 0                              & 8                                                  & 16                                                  & \textbf{24}                                                     & 28                                                     & 28                                                    & 28                                                      & 32                                                     \\ \cline{2-2} \cline{7-14} 
		& fc2                          &                                                                         &                               &                               &                               & 0                              & 8                                                  & 16                                                  & \textbf{4}                                                      & 4                                                      & 8                                                     & 16                                                      & 8                                                      \\ \cline{2-2} \cline{7-14} 
		\multirow{-3}{*}{\begin{tabular}[c]{@{}c@{}}$N_w$\end{tabular}} & fc3                          & \multirow{-3}{*}{-}                                                     & \multirow{-3}{*}{-}           & \multirow{-3}{*}{-}           & \multirow{-3}{*}{-}           & 0                              & 8                                                  & 0                                                   & \textbf{0}                                                      & 0                                                      & 0                                                     & 0                                                       & 0                                                      \\ \hline
		\multicolumn{2}{|c|}{\begin{tabular}[c]{@{}c@{}}Total \\memory \\ storage\end{tabular}}                                                                       & 1070 KB                                                 & 88 KB                          & -                          & 27 KB                          & 52.5 KB                            & 13 KB                                               & 7 KB                                               &  \textbf{5.5 KB}                                                 & 4.9 KB                                                   & 6.5 KB                                                 & 6.1 KB                                                   & 5.5 KB                                                \\ \hline
		
		\multicolumn{2}{|c|}{\begin{tabular}[c]{@{}c@{}} Memory \\ reduction \\ rate\end{tabular}}                                                                       & 1 $\times$                                                                     & 12$\times$                           & 32$\times$                          & 40$\times$                           & 20.4$\times$                             & 82.3$\times$                                              & 152.9$\times$                                               &  \textbf{194.5$\times$}                                                 & 218.4$\times$                                                    & 164.6$\times$                                                  & 175.4$\times$                                                   & 194.5$\times$                                                 \\ \hline
		\multicolumn{2}{|c|}{Accuracy}                                                                                                                        & {\color[HTML]{9A0000} 98.4\%}                                           & {\color[HTML]{9A0000} 98.4\%} & {\color[HTML]{9A0000} 98.4\%} & {\color[HTML]{9A0000} 98.4\%} & {\color[HTML]{9A0000} 97.53\%} & {\color[HTML]{9A0000} 97.49\%}                     & {\color[HTML]{9A0000} 97\%}                         & {\color[HTML]{9A0000}  \textbf{96.44\%}}                         & {\color[HTML]{9A0000} 94.48\%}                         & {\color[HTML]{9A0000} 97\%}                           & {\color[HTML]{9A0000} 96.6\%}                           & {\color[HTML]{9A0000} 96.32\%}                            \\ \hline
	\end{tabular}
\end{table*}
\section{Experiments} \label{ex:sec}

\subsection{MNIST dataset }
The proposed  approximation  algorithm was tested using MATLAB   in both inference and training phases for the   MNIST image dataset.   MNIST image dataset is a large handwritten digit (0-9) which include 50,000 and 10,000 training and test images, respectively.  The number of weights  can be reduced more by increasing the number of weights inside the linear interval.  
\subsubsection{FC architecture }
Applying the proposed approximation algorithm in the inference reach a  31.78\% accuracy for 3 weights inside the linear interval which is a very low  accuracy  compared with the case without approximation.   
Moreover, using 3 weights or 4 weights or 8 weights  inside the linear interval in the training did not affect the accuracy where an accuracy of 97.53\%, 97.53\%,  and 97.49\%  are achieved for the method without approximation, 3 weights, 4 weights,  and 8 weights  inside the linear interval   respectively. 

%
%

%

% \usepackage{multirow}

Table \ref{Tab11} presents a comparison between the proposed methods and previous works \cite{han2015deep,han2015learning} P, P+Q and P+Q+H ( P: pruning, Q:quantization, H:Huffman) in-terms of   accuracy, reduction rate and number of parameters.  Pruning prunes the unimportant connections to reduce the number of parameters. Quantization then reduces the number of bits that represent each connection from 32-bits to 5-bits. The quantization is applied  to enforce weight sharing. For the proposed methods, one case (without) is introduced for a reduced version of LeNet 300-100 which is built using 64 and 32 neurons instead of 300 and 100 neurons.  Another  7 cases are given for the proposed linear and quadratic methods, where the proposed methods is applied on the reduced version LeNet 64-32. \color{black} In Table \ref{Tab11}, all the weights in FC layers are represented with 8 bits, which does not affect the accuracy. The reduced version LeNet 64-32 saves network storage by a factor of $20.4\times$, which uses memory storage of 52.5 KB compare with 1070 KB for the original LeNet  300-100. 
Applying the proposed linear method can further save the storage memory by a factor of  $82.3\times$,  $152.9\times$,   $192.5\times$   with accuracy  97\%, 96.44\% and 94.48\% respectively.  The memory storage can be reduced more by reducing the number of bits for each weight. For example, using 6 bits instead of 8 bits reduces the memory from 5.5 KB to 4.125 KB for case 3 at Table \ref{Tab11}. In which, it can reduce the memory storage by a factor of  259.3$\times$ instead of 194.5$\times$.     \color{black}
Similar to the linear method,  the total number of parameters can be reduced by $164.6\times$  and $ 175.4.3\times$ using quadratic approximation method with 97\% and 96.6\% respectively. 
As can be seen, the accuracy decreases as the number of weights inside the linear interval increase. On the other hand, the reduction rate is increased with increasing the number of weights inside the linear interval.  The reduction rate for the weights can be computed as follows:
\begin{equation}
	reduction~rate~(r)=\frac{n}{\frac{2n}{N_w}}=\frac{N_w}{2}
\end{equation}
where n is the total number of weights and $N_w$ depicts the number of weights inside the linear interval.

\subsubsection{CNN architecture}
The proposed algorithm is applied for  MNIST CNN with the following properties:
\begin{itemize}
	\item  Input (map size: $28\times28$)
	\item Convolution with $6$ kernels of size $5\times5 + sigmoid (24\times24\times6)$.
	\item Average pooling with $2\times2$ kernel $(12\times12\times6)$.
	\item Convolution with $6\times12$ kernels of size $5\times5 + sigmoid (8\times8\times12)$
	\item Average pooling with $2\times2$ kernel $(4\times4\times12) + vectorization (192\times1)$
	\item Fully connection + sigmoid $=> $ output $(10\times1)$
\end{itemize}

\begin{table*}[]
	\centering
	\caption {\color{black}Comparison with LeNet 5 (CNN approach) and  \cite{han2015deep,han2015learning} P: pruning, Q:quantization, H:Huffman coding, $N_p$  is the total number of parameters and $N_w$ depicts the number of weights inside the linear/quadratic interval. Each weight needs 8-bits fixed point for proposed method. The linear and quadratic approximation are done during the training. \color{black}  }
	\label{Tab101}
	\begin{tabular}{|c|c|c|c|c|c|c|c|c|c|c|c|c|}
		\hline
		\multicolumn{2}{|c|}{}                                                          &                                                                         &                            &                          &                          & \multicolumn{7}{c|}{Proposed CNN}                                                                                                                                                                                               \\ \cline{7-13} 
		\multicolumn{2}{|c|}{}                                                          &                                                                         &                            &                          &                          &\begin{tabular}[c]{@{}c@{}} Without \\ approximation  \end{tabular}                    & \multicolumn{4}{c|}{ \begin{tabular}[c]{@{}c@{}} Linear \\approximation  \end{tabular}}                                                                                                    & \multicolumn{2}{c|}{ \begin{tabular}[c]{@{}c@{}} Quadratic\\ approximation  \end{tabular}}                                 \\ \cline{7-13} 
		\multicolumn{2}{|c|}{\multirow{-3}{*}{Method}}                                  & \multirow{-3}{*}{\begin{tabular}[c]{@{}c@{}}Ref\\ LeNet 5\end{tabular}} & \multirow{-3}{*}{\begin{tabular}[c]{@{}c@{}}P \\ \cite{han2015deep} \\ \cite{han2015learning}\end{tabular}}        & \multirow{-3}{*}{\begin{tabular}[c]{@{}c@{}}P+Q \\ \cite{han2015deep}\end{tabular}}    & \multirow{-3}{*}{ \begin{tabular}[c]{@{}c@{}} P+Q+H \\ \cite{han2015deep}\end{tabular}}  & Case 0                        & Case 1                        & Case 2                         & Case 3                        & Case 4                        & Case 5                        & Case 6                         \\ \hline
		\multicolumn{2}{|c|}{$N_p$}                                                      & {\color[HTML]{9A0000} 431k}                                             & {\color[HTML]{9A0000} 22k} & {\color[HTML]{9A0000} -} & {\color[HTML]{9A0000} -} & {\color[HTML]{9A0000} 3.87k}  & {\color[HTML]{9A0000} 1.42k}  & {\color[HTML]{9A0000} 0.9k}    & {\color[HTML]{9A0000} 0.84k}  & {\color[HTML]{9A0000} 0.82k}  & {\color[HTML]{9A0000} 1.2k}   & {\color[HTML]{9A0000} 0.294k}  \\ \hline
		& Conv                         &                                                                         &                            &                          &                          & 0                             & 5                             & 5                              & 5                             & 5                             & 5                             & 25                             \\ \cline{2-2} \cline{7-13} 
		\multirow{-2}{*}{$N_w$}                           & FC                           & \multirow{-2}{*}{-}                                                     & \multirow{-2}{*}{-}        & \multirow{-2}{*}{-}      & \multirow{-2}{*}{-}      & 0                             & 6                             & 32                             & 64                            & 96                            & 192                           & 96                             \\ \hline
		\multicolumn{2}{|c|}{\begin{tabular}[c]{@{}c@{}}Total \\ memory \\ storage\end{tabular}} & 1720 KB                                                                     & 88 KB                         & -                      & 44 KB                      & 3.87 KB                           & 1.42 KB                       & 0.9 KB                          & 0.84 KB                           & 0.82 KB                         & 1.2 KB                       & 0.294 KB                         \\ \hline
		\multicolumn{2}{|c|}{\begin{tabular}[c]{@{}c@{}}Memory \\reduction \\ rate\end{tabular}} & 1$\times$                                                                      & 12$\times$                         & 33$\times$                       & 39$\times$                       & 444.4$\times$                           & 1211.3$\times$                         & 1911.1$\times$                          & 2047.6$\times$                           & 2097.6$\times$                         & 1433.3$\times$                         & 5850.3$\times$                          \\ \hline
		
		\multicolumn{2}{|c|}{Accuracy}                                                  & \multicolumn{4}{c|}{{\color[HTML]{9A0000} 99.1\%}}                                                                                                         & {\color[HTML]{9A0000} 98.8\%} & {\color[HTML]{9A0000} 98.2\%} & {\color[HTML]{9A0000} 97.63\%} & {\color[HTML]{9A0000} 95.1\%} & {\color[HTML]{9A0000} 93.6\%} & {\color[HTML]{9A0000} 93.4\%} & {\color[HTML]{9A0000} 91.34\%} \\ \hline
	\end{tabular}
\end{table*}

%The proposed algorithm is applied in both inference and training phases for verification.
%Applying the proposed algorithm in the training phase.  
For CNNs, the proposed algorithm is applied in both  the convolution and FC layers, where each row in the filter $5\times5$ is approximated linearly as presented in Fig. \ref{fig_1}. The linear interval has a five weights. For the FC layer, the number of weights is $192\times10$. Each $192$ is divided to uniform groups where all the groups have the same number of weights. Any group of weights is approximated based on the algorithm presented in Fig. \ref{fig_4}.   
Applying the proposed algorithm in the inference shows a poor results where the testing accuracy is 48.7\%.  
Table \ref{Tab101} presents a comparison with LeNet 5 (CNN approach) and  \cite{han2015deep,han2015learning} P, P+Q and P+Q+H in-terms of   accuracy, reduction rate and number of parameters.  Without approximation case is a reduced version LeNet 5, which  is the proposed MNIST CNN.  
As can be seen, the proposed method achieved the best parameters reduction rate, which achieved 1211.3$\times$ compared with 39$\times$ for  \cite{han2015deep} with drop by only  0.9\% from the accuracy.  \color{black}All the results in  \cite{han2015deep,han2015learning} are achieved using 8-bits fixed point for the representation of each weight. \color{black}

\begin{figure}[t]
	\centering
	\subfloat[]{\includegraphics[width=0.50\linewidth]{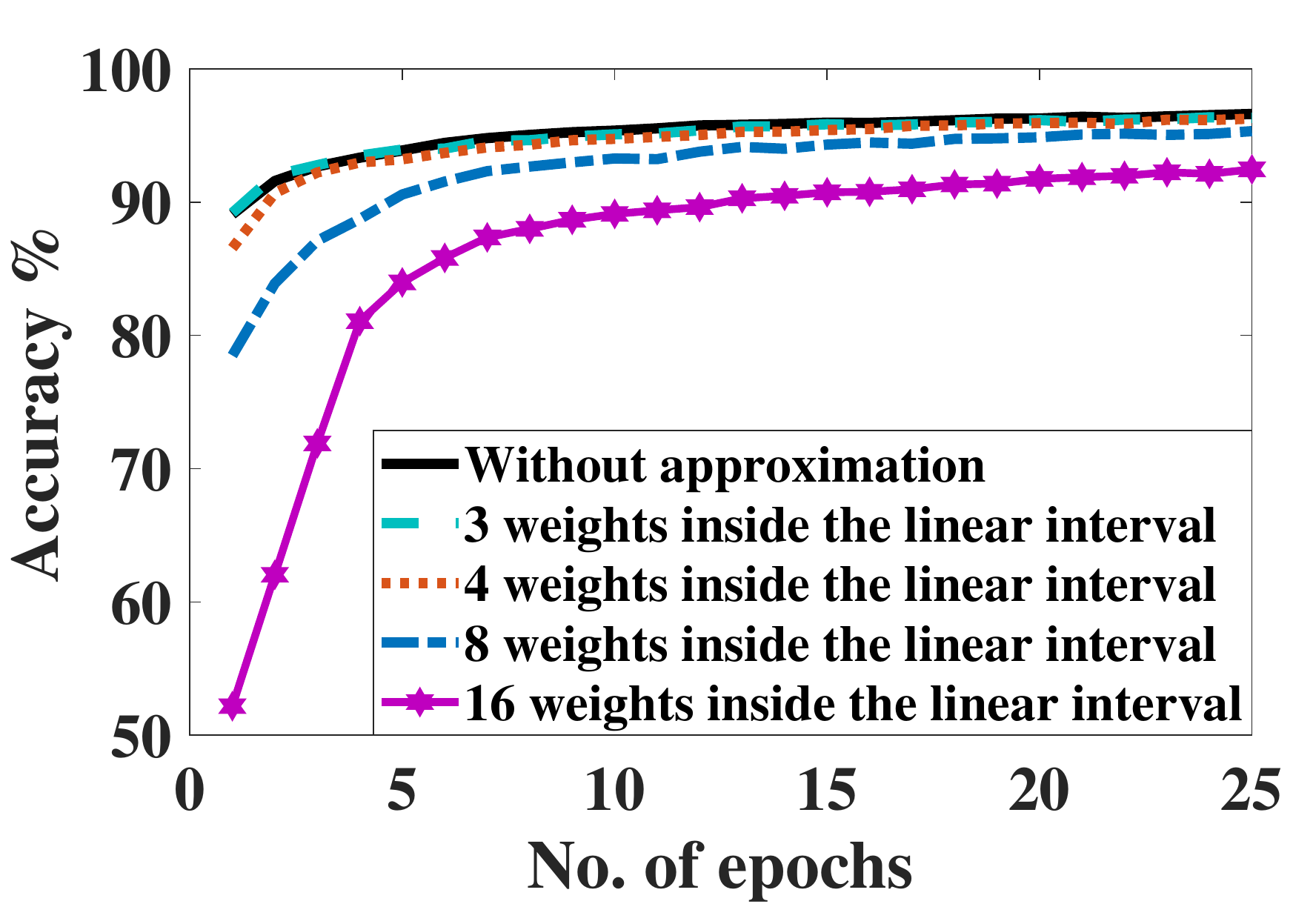}
		\label{fig_all}}
	\subfloat[]{\includegraphics[width=0.48\linewidth]{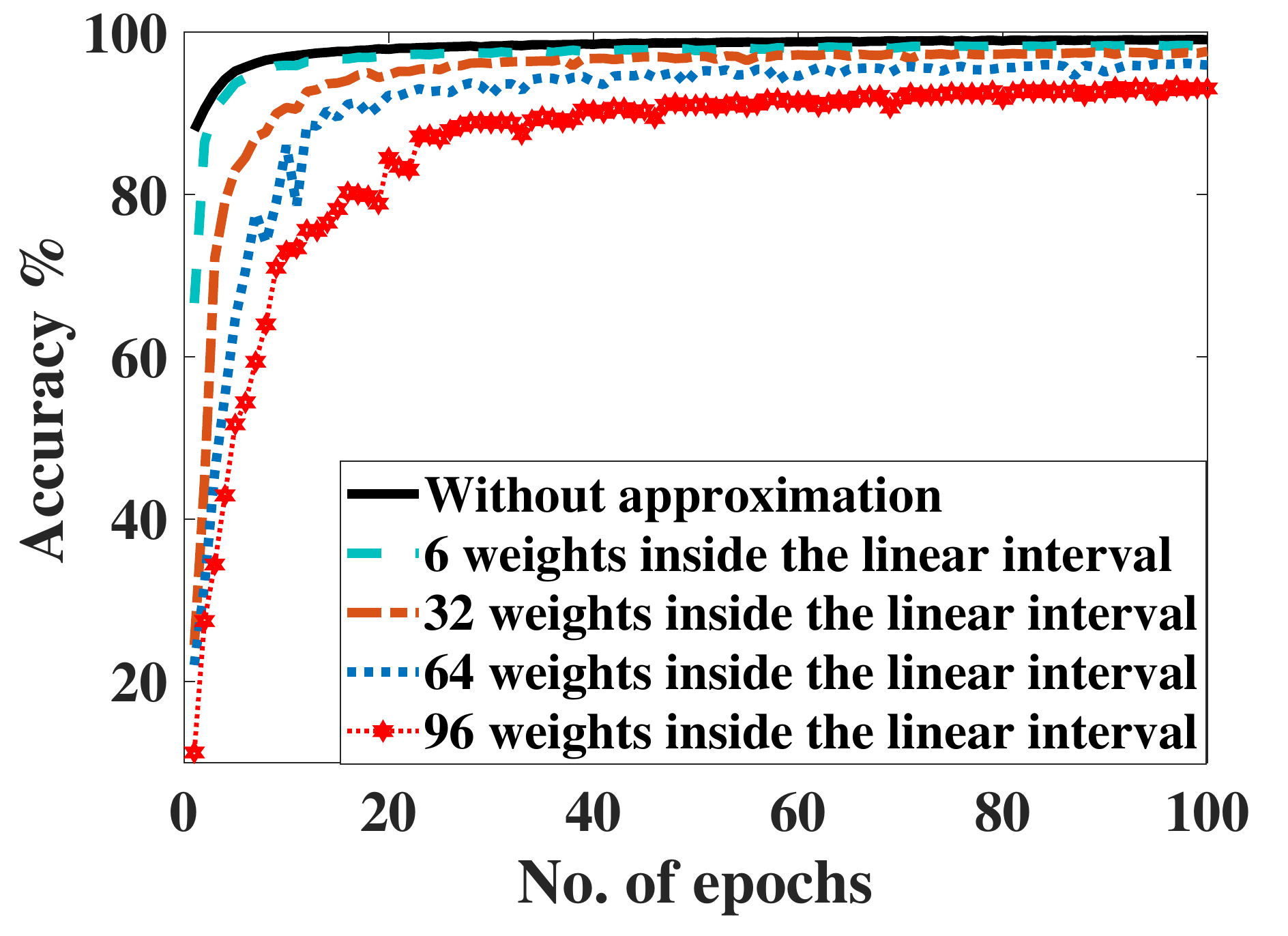}
		\label{acc_cnn}}
	\caption{ Testing accuracy vs  number of epochs for different number of weights inside the linear interval (a)  MNIST FC with two hidden layers (64 and 32). (b)  MNIST CNN, all cases use 5 weights inside the linear interval at the convolution layer. 6, 32, 64 and 96   weights are used inside the linear interval for the FC layer. }
	\label{acc_cnn2}
\end{figure}

Figure \ref{fig_all} shows the test accuracy vs number of epochs for a neural network based on  FC  with two hidden layers (64 and 32). The accuracy is computed for the method with and without linear approximation method (method presented in Fig. \ref{fig_4}).
Figure \ref{acc_cnn}  shows the  testing accuracy vs number of epochs for different number of weights inside the linear interval for CNN architecture. The proposed algorithm is applied for 5 weights inside the linear interval for all the convolution layers where each filter row is approximated linearly. For the FC, the proposed algorithm is applied for  6, 32, 64 and 96 weights inside the linear interval.   As can be seen from the plot, the drop in testing accuracy is not significant for most of the approximation cases.

%Figure \ref{fig_linea_2nd_acc} presents   a comparison between linear and quadratic approximation methods for different number of weights inside the approximation interval  for  MNIST  CNN architecture.  Figure \ref{fig_lin1}  shows testing accuracy vs  number of epochs 25 and 96 weights inside the approximation interval for each convolution and FC layer respectively. A 5 and 192 weights inside the approximation interval  are presented in Fig. \ref{fig_lin2} for each convolution and FC layer respectively. As can be seen, the accuracy is enhanced using quadratic approximation compared with linear method which  the quadratic  and linear methods achieved 93.4\% and 75.59\% accuracy for the case presented in Fig. \ref{fig_lin2} respectively. 

\begin{table}[t]
	\centering
	\caption {\color{black}Testing accuracy and weights reduction rate for different number of weights inside the linear interval $N_w$  for CIFAR 10. $N_P$ is the number of parameters. Each weight needs 8-bits fixed point for proposed method. The training is applied for 20 epochs.\color{black}  }
	\label{Tab111}
	\begin{tabular}{|c|c|c|c|c|c|c|}
		\hline
		\multicolumn{2}{|c|}{Approximation}                                               & baseline   & \begin{tabular}[c]{@{}c@{}} During \\inference\end{tabular} & \multicolumn{3}{c|}{During the training} \\ \hline
		\multirow{2}{*}{$N_w$}                                                     & convs & 0       & 5                & 5            & 5           & 5           \\ \cline{2-7} 
		& fcs   & 0       & 8                & 8            & 16          & 32          \\ \hline
		\multirow{3}{*}{$N_p$}                                                     & convs & 256.8k  & 102.8k           & 102.8k       & 102.8k      & 102.8k      \\ \cline{2-7} 
		& fcs   & 263.4   & 65.8k            & 65.8k        & 32.9k       & 16.4k       \\ \cline{2-7} 
		& Total & 520.2k  & 158.6            & 158.6        & 135.7k      & 119.2k      \\ \hline
		\multicolumn{2}{|c|}{\begin{tabular}[c]{@{}c@{}}Total memory \\ storage "KB" \end{tabular}}                                                    & 2080     & 158.6              & 158.6        & 135.7        & 119.2       \\ \hline
	\multicolumn{2}{|c|}{\begin{tabular}[c]{@{}c@{}}Memory \\reduction rate\end{tabular}}  & 1$\times$       & 13.1$\times$              & 13.1$\times$          & 15.3$\times$         & 17.4$\times$         \\  \hline
		\multicolumn{2}{|c|}{Accuracy}                                                    & 76.2    & 15.3             & 75\%         & 73.6\%      & 72.2\%      \\ \hline
	\end{tabular}
\end{table}

\subsection{ CIFAR 10 dataset   }

To demonstrate the effectiveness of our proposed linear approximation algorithm we performed more experiments on the CIFAR-10 dataset. CIFAR-10 consists of 50,000 training images and 10,000 testing images of size $32\times32$ from 10 classes. The architecture used for CIFAR 10   with the following details
\begin{itemize}
	\item  Input (map size: $32\times32$)
	\item Convolution with $32$ kernels of size $5\times5 + relu (32\times32\times32)$.
	\item Average pooling with $2\times2$ kernel $(16\times16\times32)$.
	\item Convolution with $32\times64$ kernels of size $5\times5 + relu (16\times16\times64)$.
	\item Average pooling with $2\times2$ kernel $(8\times8\times64)$.
	\item Convolution with $64\times128$ kernels of size $5\times5 + relu (8\times8\times128)$.
	\item Average pooling with $2\times2$ kernel $(4\times4\times128) + vectorization (2048\times1)$.
	\item Fully connection + relu $=> $ output $(128\times1)$.
	\item Fully connection + relu $=> $ output $(10\times1)$.
	\item softmax.
\end{itemize}
\begin{figure*}[t]
	\centering
	{\includegraphics[width=0.9\linewidth]{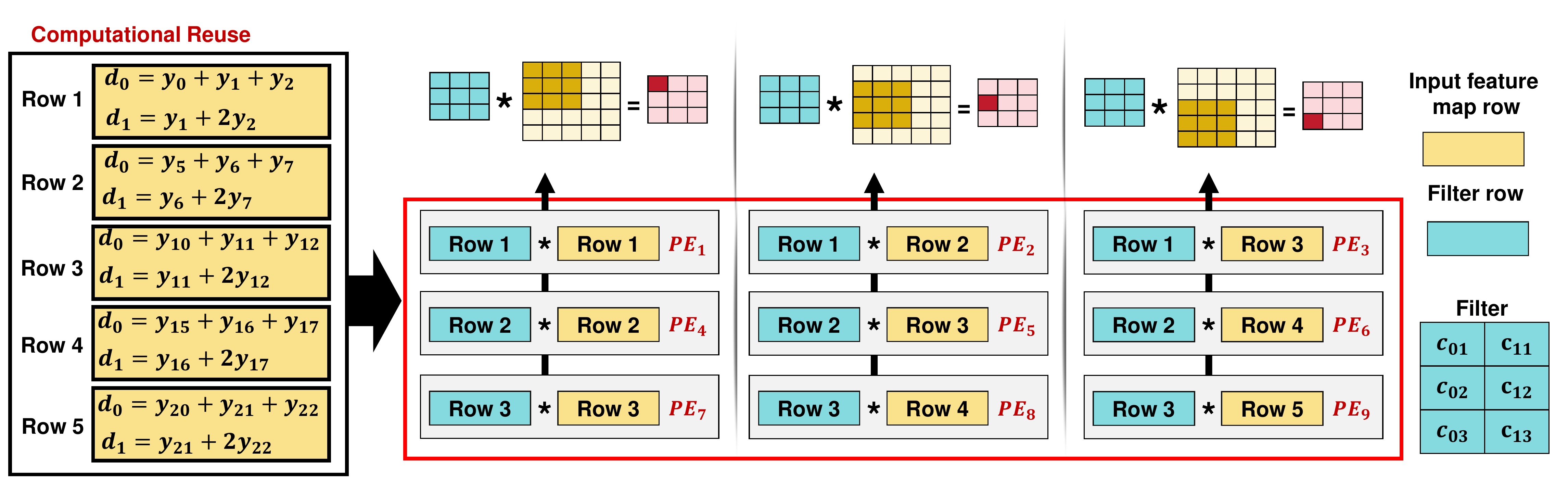}
		\caption{Example for the proposed CNNs architecture, where filter reuse, input feature map reuse and computational reuse are achieved. Each PE requires 2 multipliers and one adder, for example $PE_1$ perform the operation $d_0\times c_{01} + d_1\times c_{11}$.   }
		\label{fig_2}}
\end{figure*}
Our target is to study the effect of our proposed algorithm on the accuracy.  The approximation was applied on every batch throughout the training period with learning rate 0.001. 
Table \ref{Tab111} shows the accuracy, number of parameters and reduction rate for different number of weights inside the linear interval for CIFAR 10 dataset. Applying the linear approximation algorithm  in the inference shows a poor results where the testing accuracy is 15.3\%. 
The proposed algorithm is applied in the training for 20 epochs.  A reduction rate of $13.1\times$ is achieved   by applying the linear approximation  in the training.


\begin{figure*}[]
	\centering
	{\includegraphics[width=0.9\linewidth]{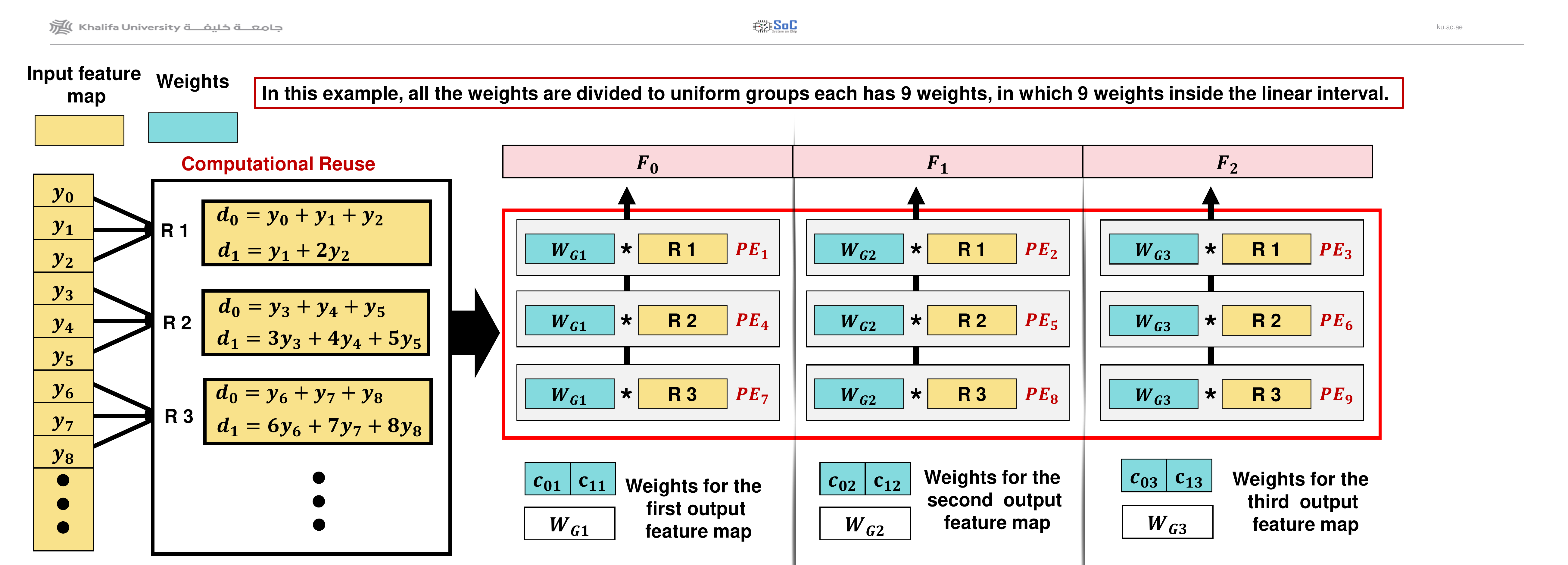}
		\caption{Example for the proposed FC architecture, where  input feature map reuse and computational reuse are achieved Each PE requires 2 multipliers and one adder, for example $PE_1$ perform the operation $d_0\times c_{01} + d_1\times c_{11}$.   }
		\label{fig_6}}
\end{figure*}

\section{Opportunity for  Data reuse and Computation Reuse} \label{reuse}
In Fig. \ref{fig_1} equation (2), the convolution of input feature map row 1 $(y_0,y_1,y_2)$ and filter row 1 $w_0,w_1,w_2$ need the following computations $ d_0=y_0+y_1+y_2$ and $d_1=y_1+2y_2$.   Based on the data reuse technique, those computations could be reused inside the PE array processor. Figure \ref{fig_2} presents an example to generate the output partial sums for CNN architecture, in which filter reuse, input feature map reuse, and computational reuse methods are used. The computational reuse block on the left-hand side of Fig. \ref{fig_2} contains the operations needed for each input feature map rows. These operations are reused diagonally for multiple PEs processor. The partial sums are accumulated vertically for the PEs to generate the first column of the output feature map. To produce the second column of the output, all the selected elements at each row of input feature maps are shifted right by one column. The number of PE columns and rows equal the number of output feature map rows and the number of filter rows respectively. The amount of data reuse and computational reuse in the DNN layer (input feature map, filter weight, and output feature map) is a function of the layer shape and size.

Figure \ref{fig_6} gives an example of producing the output partial sum for FC architecture. On the  left is the  input feature map $y_0:y_n$, which drives the computational reuse block to produce the sub-computation needed in the PEs array.  In this  example, the number of weights inside the linear interval is 9 which means any 9 weights shared the same value.The computational reuse block contains several sub-blocks ($R_1, R_2$ and $R_3$ for each 9  $ifmap$) which includes the the computation required for  PEs. Fully connected layers are typically applied on the features extracted from the convolution layers for classification purposes. The FC layer also applies filter on the $ifmap$ with kernel of the same size as the $ifmap$.
Hence, it does not 
have the weight sharing feature like convolution layers. Based on the proposed algorithm of FC structure presented in Fig. \ref{fig_4}, a repetition can be achieved for the FC weights. Taking advantage of repeated weights a weight reuse and computation reuse can be achieved as shown in Fig. \ref{fig_6}. The weights are reused vertically and the sub-computations are reused horizontally.    

The total number of adders and multipliers for the proposed CNN architecture based linear approximation method are computed as follows: 
\begin{equation} \label{eq:1}
	N_{add.mult}=N_{1}\times N_{2}+ 3 \times N_{3} 
	%N_{add.mult}=N_{ifmap.row} N_{add.ifmap.row}+3  \left( 1+ N_{PE.mult} \right)
\end{equation}
where $N_{add.mult}$ depicts the total number of adders and multipliers, $N_1$ is the number of $ifmap$ rows (or groups), $N_2$ represents the number of adders needed for each $ifmap$ row (or group) and $N_3$ is the total number of the PEs.  and $"3"$ means one adder + two multipliers inside each PE for proposed architecture.

\begin{figure}[t]
	\centering
	{\subfloat[]{\includegraphics[width=0.49\linewidth]{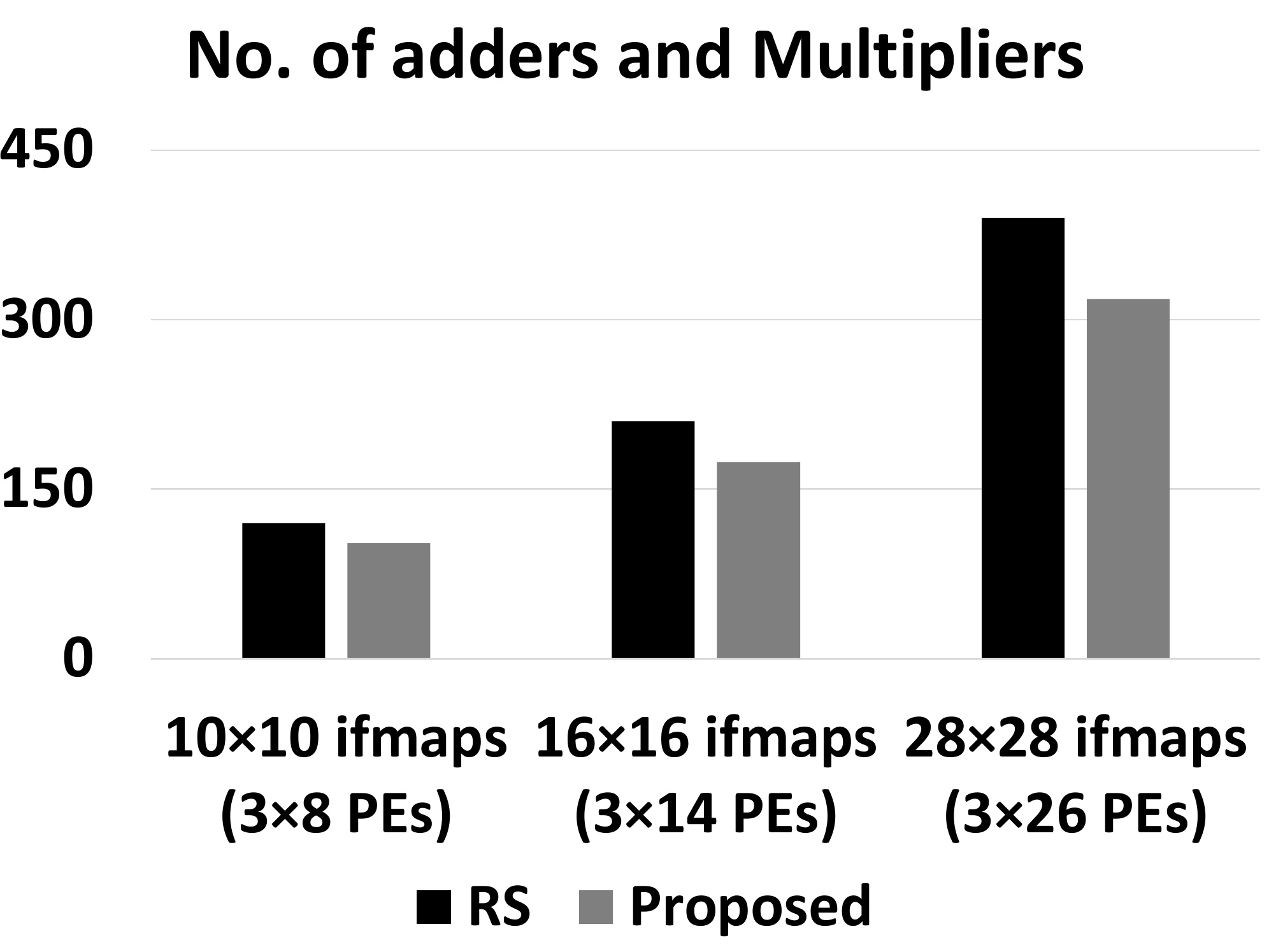}
			\label{fig_8}}
		\subfloat[]{\includegraphics[width=0.49\linewidth]{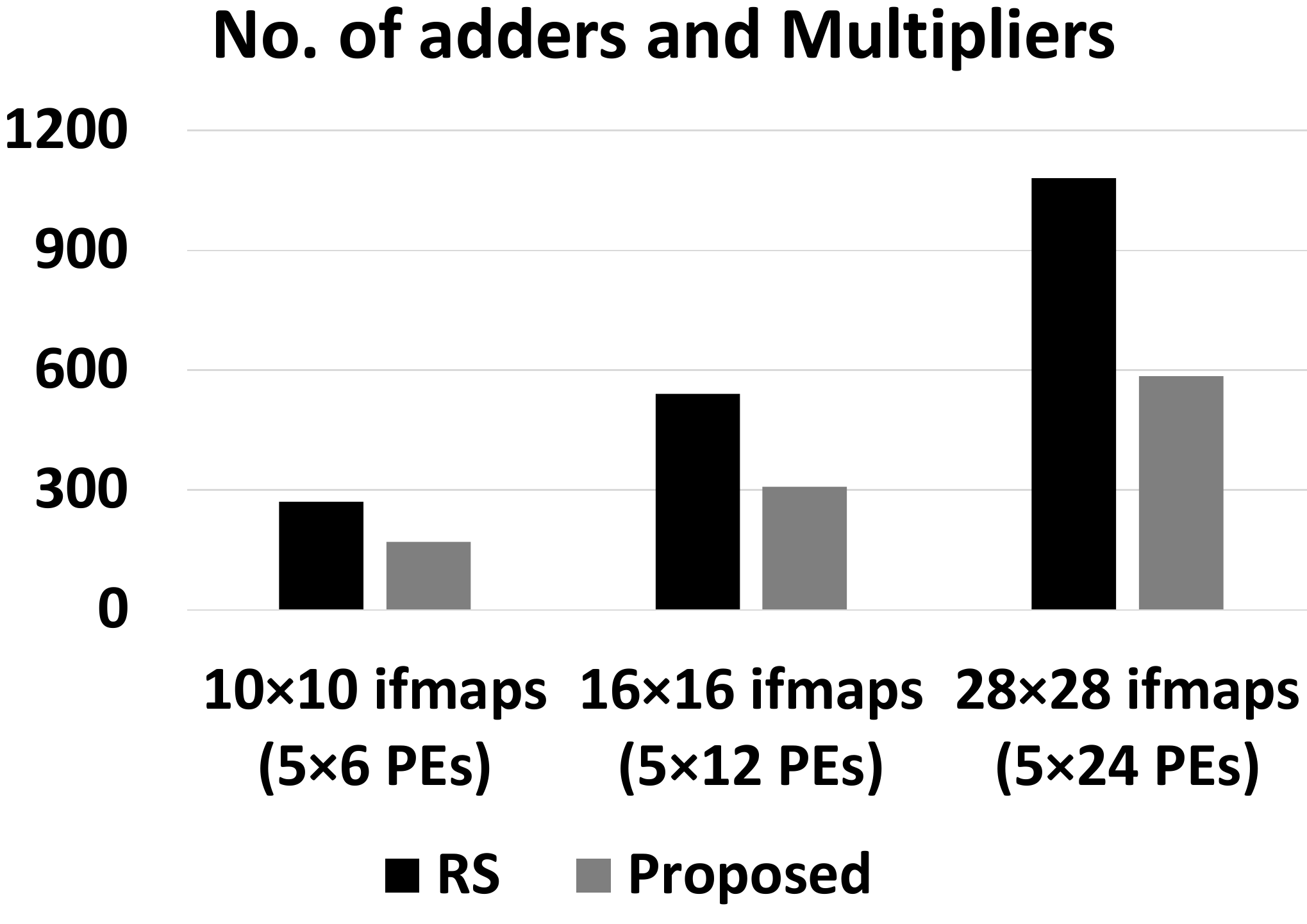}
			\label{fig_9}}
		\caption{Number of adders and multipliers for different feature maps ($ifmap$) for proposed  architecture and RS dataflow for convolution layer architecture, (a) the number needed for one filter $(3\times3)$ and (b)  the number needed for one filters $(5 \times 5)$.   }
		\label{fig_8_9}}
\end{figure}

For  RS dataflow, the total number of adders and multipliers are $N_3\times N_4$, where $N_4$ depicts the number of adders and multipliers inside each PE for RS dataflow. 
An example of (filter $3 \times 3$ and $ifmap~ 10 \times 10$),  $N_1=10$, $N_2=3$, $N_3=3\times 8$ and $N_4=5$. The total number of adders and multipliers are $102$ and $120$ respectively.  For proposed architecture (filter $3 \times 3$), each input feature map row demands 3 adders as given in Fig. \ref{fig_2}, each PE uses one adder and two multipliers to perform $c_0 \times d_0 + c_1 \times d_1$.
For $ifmap~10 \times 10$ and one filter $3 \times 3$, an output feature map $3\times8$ is generated.
Figure \ref{fig_8_9} shows a comparison based on the total number of adders and multipliers for different input feature maps ($ifmap$) for convolution layer architecture.
Figure \ref{fig_8_9} (a) and (b) show the   total number of adders and multipliers for using a $3\times3$ filter and $5\times5$ filter respectively. As can be seen, the total number of adders and multipliers are reduced more when using $(5\times5)$. For input feature map $28\times28$ and PE array of $5\times24$, the total number of adders and multipliers are 584 and 1080 for proposed architecture and RS dataflow receptively. Compare to state of the art RS the proposed architecture saved  54\% of number of adders and multipliers.

Table \ref{Tab9} shows the total number of adders and multipliers required to implement the inference for  LeNet 300-100 based different cases.  
The number of adders and multipliers needed to implement  LeNet 300-100 is 531990. For the reduced version FC (64-32), needs 104982 adders and multipliers.  Using the proposed method, requires 11902 adders and multipliers for case 4. In which, the proposed method achieved a reduction rate of $40.4\times$ and $44.7\times$ in the number of adders and multipliers for case 3 and 4 receptively. Case 0, 3 and 4 were the cases introduced in Table \ref{Tab11}.

\begin{table}[]
	\centering
	\caption {Comparison based  computations reduction rate, total number of adders and multipliers for LeNet 300-100, case 0, case 3 and case 4 are the cases presented in Table \ref{Tab11}.   }
	\label{Tab9}
	\begin{tabular}{|c|c|c|c|c|}
		\hline
		\multirow{2}{*}{}                                                          & \multirow{2}{*}{LeNet 300-100} & FC 64-32 & \multicolumn{2}{c|}{\begin{tabular}[c]{@{}c@{}}Proposed linear \\ approximation\end{tabular}} \\ \cline{3-5} 
		&                                & Case 0   & Case 3                                        & Case 4                                        \\ \hline
		\begin{tabular}[c]{@{}c@{}}No. of adder \\ and  multipliers\end{tabular} & 531990                         & 104982   & 13156                                         & 11902                                          \\ \hline
		
		\begin{tabular}[c]{@{}c@{}}Computations \\ reduction rate\end{tabular}                                                           & 0                              & 5$\times$       & $40.4\times$                                          & $44.7\times$                                         \\ \hline
	\end{tabular}
\end{table}

\section{Conclusions}\label{sec:conclusion}

We have introduced a technique to improve the energy efficiency and memory storage needed for neural networks with a small effect on the accuracy. Our method, motivated in part by presenting a new DNN approximation algorithm that reduces the number of weights and exploits weight repetition to reduce the number of needed multipliers and adders. 
Linear and quadratic approximation methods were used to approximate the DNN weights during the training phase.
The proposed technique was evaluated with LeNet 300-100 and LeNet 5 models using MNIST and CIFAR 10 dataset. It has reduced the number of adders and multipliers required by a factor of $40.4\times$ for LeNet 300-100.
The proposed method leads to an inference implementation with smaller memory storage and  a reduced number of computations which makes DNNs implementation more feasible on edge devices. 
Furthermore, the proposed method allows deployment of complex neural networks in mobile platforms with constrained download bandwidth, strict power budget and limited application size.

%\section*{Acknowledgment}

%\section{Acknowledgment}
%This publication is based upon work supported by System-On-Chip center at  the Khalifa University of Science and Technology under Award No. [RC2-2018-020], in addition to 
%Dubai Electronic Security Center (No. 8434000322-Energy Efficient Secure IoT Hardware for Smart Cities). 

\bibliographystyle{ieeetran}
\bibliography{bibfile}

\end{document}